# Sleep Arousal Detection from Polysomnography using the Scattering Transform and Recurrent Neural Networks

Philip Warrick[1], Masun Nabhan Homsi[2]

[1]PeriGen. Inc., Montreal, Canada
[2] Simon Bolivar University, Caracas, Venezuela

*Abstract*: Sleep disorders are implicated in a growing number of health problems. In this paper, we present a signal-processing/machine learning approach to detecting arousals in the multi-channel polysomnographic recordings of the Physionet/CinC Challenge2018 dataset.

*Methods:* Our network architecture consists of two components. Inputs were presented to a Scattering Transform (ST) representation layer which fed a recurrent neural network for sequence learning using three layers of Long Short-Term Memory (LSTM). The STs were calculated for each signal with downsampling parameters chosen to give approximately 1 s time resolution, resulting in an eighteen-fold data reduction. The LSTM layers then operated at this downsampled rate.

*Results:* The proposed approach detected arousal regions on the 10% random sample of the hidden test set with an AUROC of 88.0% and an AUPRC of 42.1%.

## 1. Introduction

Polysomnography (PSG) is a common medical test mainly employed to diagnose a variety of sleep disorders such as sleep apnea, REM sleep behaviour disorder, restless leg syndrome and so on. The components of PSG include electromyography (EMG), electrooculography (EOG), electroencephalography (EEG), electrocardiology (ECG), oxygen saturation (SaO2) and respiratory airflow (AIRFLOW). Annotating the sleep stages of the PSG recordings is a slow demanding task usually carried out manually by a sleep expert. Sleep arousal interruptions last usually from 3 to 15 seconds and can occur spontaneously or as a result of sleep-disordered breathing or other sleep disorders. Left untreated, sleep disorders can increase the risk for a multiplicity of health problems, including heart disease. Having a tool to automatically annotate PSG recordings could help clinicians to quantify and choose appropriate treatments.

Approaches to automatic identification of arousal regions from physiological recordings have used a single data modality such as ECG [5] or EEG [1], or combinations (EEG, EOG and EMG [3], EEG, AIRFLOW and EOG [8], or ECG and SpO2 [7]). Traditionally, features are extracted from the time or frequency domain followed by a machine learning classifier.

In this paper, we explore the use of the Scattering Transform (ST) for feature extraction along with deep Recurrent Neural Network (RNN), especially Long Short-term Memory (LSTM) networks to predict arousal regions in 13 PSG recordings of the Physionet/CinC Challenge2018 dataset [4]. To best of our knowledge, this is the first study that investigates how the combination of these techniques is efficient on sleep arousal recognition using multimodal time series data.

ST is a non-linear mathematical operator whose characteristics are inspired by convolutional networks [6]. ST has been used successfully in many classification tasks, including musical genre, stationary textures and small digits. ST uses multiple layers of wavelet transforms, along with complex modulus operations and low-pass filter averaging. Its main objective is to provide a representation which is invariant to translation and stable to small time-warping deformations.

The remainder of this paper is organized as follows. Our methodology is presented in section 2. We perform extensive experiments and present our results and evaluations in section 3. Finally, conclusions and future directions are outlined in section 4.

## 2. Methods

Fig. 1 depicts the methodology used to detect target arousal regions in PSG recordings. The method consists of three phases: dataset preparation, classification and evaluation. These phases are described below:

### 2.1. Dataset Preparation Phase

The PhysioNet/CinC Challenge 2018 dataset [4] was first partitioned such that 10% of the data was set aside as a Held-Out Test (HOT) set, with the other 90% was used for ten-fold cross-validation, each fold being partitioned as training (90%), validation (10%) and testing (10%).

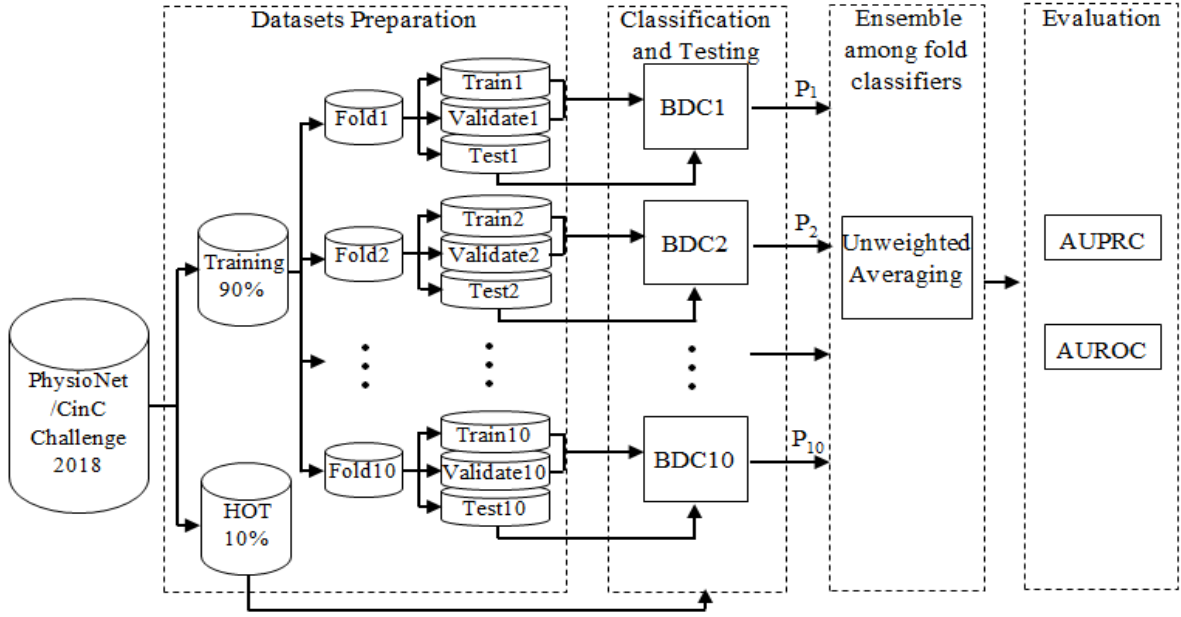

Figure 1. Block diagram of the proposed methodology.

## 2.2. Classification Phase

The arousal detection task from the 13 channel PSG recordings was carried out through two stages: building the base deep classifier and building the ensemble classifiers.

### 2.2.1. Building the Base Deep Classifier

Ten base deep classifiers (BDC) were trained, validated and tested separately using 10-fold cross validation. The BDC architecture consists of two main components. The proposed architecture is shown in Fig. 2 and is outlined below.

i) *Representation Learning*. This component used the ST algorithm to provide a multi-layer representation for an input raw signal X. ST translation invariance refers a signal and its shifted version having the same feature-space representation, while ST stability means that a signal and its slightly time warped version are mapped closely in the same feature space. The stability is achieved by applying the logarithmic filter-bank to a stretched signal to alleviate the high frequencies.

The wavelet transform of X is computed to obtain the scattering coefficients of the first layer: $|X * \Psi_{j_1,\lambda_1}| * \Phi_J$ where $j_1$ and $\lambda_1$ denotes different scales and orientations [6]. In the frequency domain, $\Phi$ plays the role of a low-pass filter, and the $\Psi$ are band-pass filters for higher frequency bands. The deleted high-frequency contents of the signal are recovered by using another set of wavelets: $|X * \Psi_{j_1,\lambda_1}| * \Psi_{j_2,\lambda_2}$. This representation is not invariant to translations but this can be achieved by averaging its amplitude: $||X * \Psi_{j_1,\lambda_1}| * \Psi_{j_2,\lambda_2}| * \Phi_J$. The process of building invariance and recovering information is repeated to the $k$-th layer of scattering network [6]. The outcome of the network is a scattering vector, the concatenation of the coefficients of all layers up to $m$, having size $s = \sum_{k=0}^{m} p^k \binom{J}{k}$, where $p$ denotes the number of different orientations and $J$ denotes the number of scales [6].

The STs were calculated for each signal with an averaging window of 512 samples, reducing the sampling rate to $f_{out}$=200 Hz/512=0.391 Hz (chosen to give approximately 1 s time resolution). The first two orders of the ST were retained, generating 36 coefficients per signal, an eighteen-fold (512/36) data reduction.

ii) *Sequence Learning*. LSTM cells consists of four blocks: a forget gate, input gate, output gate, and the memory state [2], designed to overcome the gradient vanishing problem and learn longer term dependencies from input [9]. Our sequence learning component consists of three consecutive layers $LSTM_i$, $i \in \{1, 2, 3\}$ with 100 units each, in order to increase the length of time dependencies; the number of units was not optimized. Between each layer $LSTM_i$, there is a Batch Normalization (BN) layer that keeps values in-bounds and avoids saturation at the various processing steps. A dense layer was added on top of the third LSTM. Finally, a softmax function provided predictions.

*BDC Learning*. For each recording of length $N$, ST was applied up to two levels with a quality factor $Q$ of 1 in the scattering filter bank, resulting in an array of dimension $(N * M)$, where $N$ denotes the number of PSG signals (13) and *M*, the number of ST coefficients. This array represents the input to LSTM$_1$, whose long-tailed distribution p($X_j$) was quasi-normalized for each ST coefficient $x_{jk}$, j ϵ {1, ..., M}, and k ϵ {1, ..., N} by taking the logarithm of the scaled by its median and a constant $\mu_j$ giving the transformed coefficients $x_{jk}$'=log1p[$\mu_j$*$x_{jk}$/med($X_j$)]. To obtain the same dimensionality for each signal, all signals were zero-padded to fixed length *max_length*.

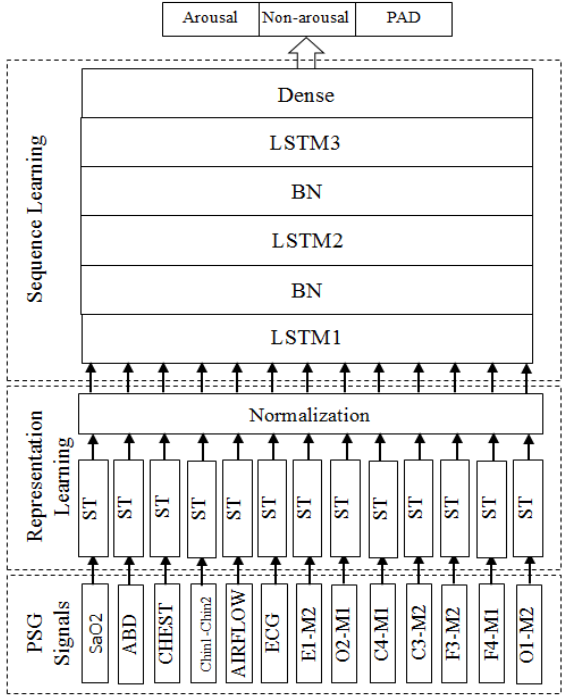

Figure 2. BDC architecture.

BDC was trained using the Root Mean Square Propagation (RMSprop) optimization method that remedies a premature drop in learning rate. The final network output *nout* is an array with the three times the dimension of the input $f_{out}$ to represent the probabilities of three labels: Arousal (2), Non-arousal (1) and PAD (0). In order to address the class imbalance between Arousal and Non-arousal regions, the loss function was weighted for targets indicating Arousal by a factor of 14 (chosen by the proportion of arousals in the training examples of a single fold) while Non-arousal and PAD targets had loss functions weighted by 1 and 0, respectively.

### 2.2.2. Building the Ensemble Classifier

Ensemble Learning refers to the combining the predictions of multiple classifiers, simulating a "committee" of decision makers. We used the unweighted averaging strategy to fuse the decision of ten BDCs. The fusion rule is given by $Y = arg_{c_j \in C} max \frac{1}{n} \sum_{i=1}^{n} P(c_j|M_i)$, where Y is the final predicted label and C represents the set of all possible labels.

The gross area under the precision-recall curve (AUPRC) and the gross area under receiver operating characteristic curve (AUROC) were used to quantify the performance of the proposed classifier.

### 3. Results and Discussions

All the experiments were performed on a PC with 32GB RAM and a 12 GB GPU. It took approximately 15 hrs to train each fold classifier. We employed the ScatNet library [10] for the ST (in the first phase of the work) and the Scattering.m library [11] (for the final phase in order to obtain an entry that could run in the time and memory budget of the Challenge server), Keras/Tensorflow for network training and the Keras CuDNNLSTM layer for optimized computing of the LSTM on GPU. The hyper-parameters of BDC were chosen experimentally and they are: 0.1 for learning rate with early stopping when the validation loss failed to improve after 50 epochs. *max_length*=13371 was set to the longest (decimated) sequence length from the training examples.

Transformed PSG signals yielded from the second layer of the scattering network of tr03-0005 recording are shown in Fig. 3, demonstrating correlation between arousal regions and transformed PSG signals.

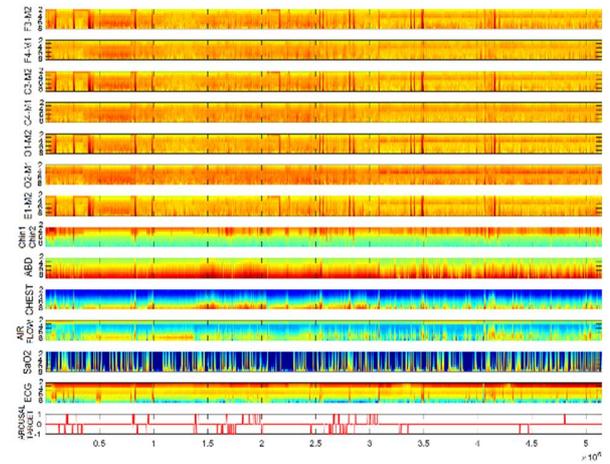

Figure 3. Arousal target (bottom red plot) and ST output for each channel (recording tr03-0005).

To estimate which were the most predictive input signals, we first trained classifiers using specific groups of inputs on a single Fold1. Table 1 shows that the All EMG classifier had the best single-mode AUPRC (19.32%) while the AIRFLOW classifier yielded the lowest (9.84%). Interestingly, using all EEG signals performed only marginally better (15.88%) than a single EEG signal (13.02%). Using all signals performed significantly better (37.54%).

Table 1. Fold1 performance of signal-group classifiers.

| Classifier | %AUROC | %AUPRC | Classifier | %AUROC | %AUPRC |
|---|---|---|---|---|---|
| All | **87.28** | **37.54** | F3-M2 | 73.66 | 13.02 |
| All EMG | 76.62 | 19.32 | ECG | 63.95 | 10.06 |
| SaO2 | 73.94 | 17.01 | AIRFLOW | 64.30 | 9.84 |
| All EEG | 77.09 | 15.88 | | | |

Then using all inputs, ten BDCs were trained and tested using 10-fold cross validation and further tested with the common HOT dataset. The test AUPRC ranged from 28.10% to 36.40%, while AUROC ranged from 80.15% to 85.64%. Afterwards, we created ensemble classifiers using increasing numbers of fold classifiers (from 2-10). Fig. 4 shows that the HOT AUPRC increased with ensemble size from 36.4% to a plateau of 43.3% at size 7.

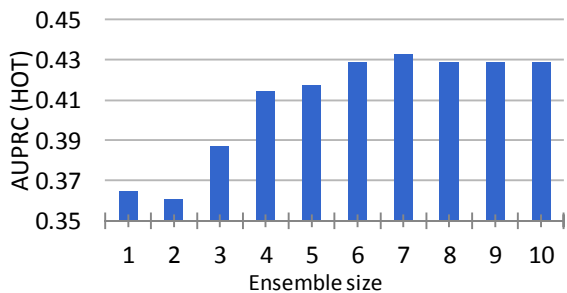

Figure 4. AUPRC vs. ensemble size on the HOT dataset.

Finally, to generate the challenge entry we trained ten classifiers new using the whole training dataset (i.e., without HOT). The per-fold results were similar to the previous experiment and on the 10% random sample of the hidden test dataset we achieved 88.0% and 42.1% for AUROC and AUPRC, respectively.

Fig. 5 compares between the annotated arousal regions in an example from Fold1 test partition, tr05-1377 recording, done manually by a sleep expert (red line) and automatically by our proposed arousal region detector (blue line). It can be seen that almost all arousal regions were successfully detected.

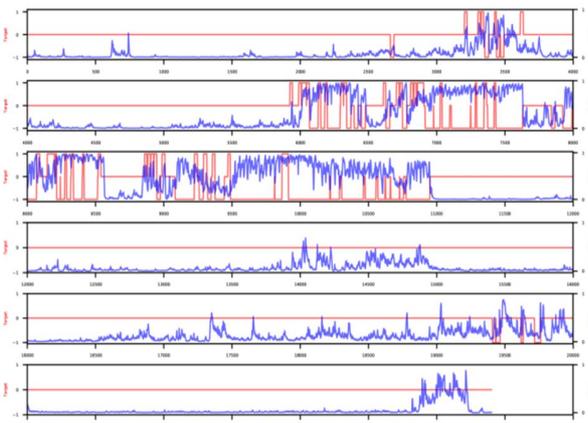

Figure 5. Automatic (blue line) vs manual annotations (red line) for a Fold1 test case, recording tr05-1377.

## 4. Conclusions

This work demonstrates a competitive approach for large scale arousal regions recognition in multimodal PSG recordings, based on scattering networks and without applying any prior feature engineering. The proposed deep architecture was capable of following the abrupt transitions in sleep biophysical signals of sleep disorder patients with a performance that was very competitive with the other participants.

## 5. Acknowledgements

The authors acknowledge PeriGen, Inc. for providing computing resources and Vincent Lostanlen, for his gracious assistance modifying his Scattering.m software to run within the Challenge2018 computing constraints.

Address for correspondence: philip.warrick@gmail.com